\documentclass[10pt,twocolumn,letterpaper]{article}

\usepackage[pagenumbers]{cvpr} 

\usepackage[ruled,linesnumbered]{algorithm2e}
\usepackage{makecell} 
\usepackage{multirow}
\usepackage{colortbl}
\usepackage{adjustbox}
\usepackage{xcolor}
\usepackage{tablefootnote}

\definecolor{cvprblue}{rgb}{0.21,0.49,0.74}
\usepackage[pagebackref,breaklinks,colorlinks,allcolors=cvprblue]{hyperref}

\def\paperID{1367} 
\def\confName{CVPR}
\def\confYear{2026}

\title{Vision Bridge Transformer at Scale}

\makeatletter
\def\@fnsymbol#1{\ensuremath{\ifcase#1\or *\or \dagger\or \ddagger\or
   \mathsection\or \mathparagraph\or \|\or **\or \dagger\dagger
   \or \ddagger\ddagger \else\@ctrerr\fi}}
\makeatother

\author{
    Zhenxiong Tan\textsuperscript{1} \quad
    Zeqing Wang\textsuperscript{1} \quad
    Xingyi Yang\textsuperscript{2,1} \quad
    Songhua Liu\textsuperscript{3,1} \quad
    Xinchao Wang\textsuperscript{1}\thanks{Corresponding Author.} \\
    \textsuperscript{1}National University of Singapore\\
    \textsuperscript{2}The Hong Kong Polytechnic University\quad
    \textsuperscript{3}Shanghai Jiao Tong University
}

\begin{document}

\twocolumn[{
\maketitle
\begin{center}
    \centering
    \vspace{-1.1cm}
    \href{https://yuanshi9815.github.io/ViBT_homepage}{\texttt{https://Yuanshi9815.github.io/ViBT\_homepage}}
    \captionsetup{type=figure}
    \vspace{0.3cm}
    \includegraphics[width=\textwidth]
    {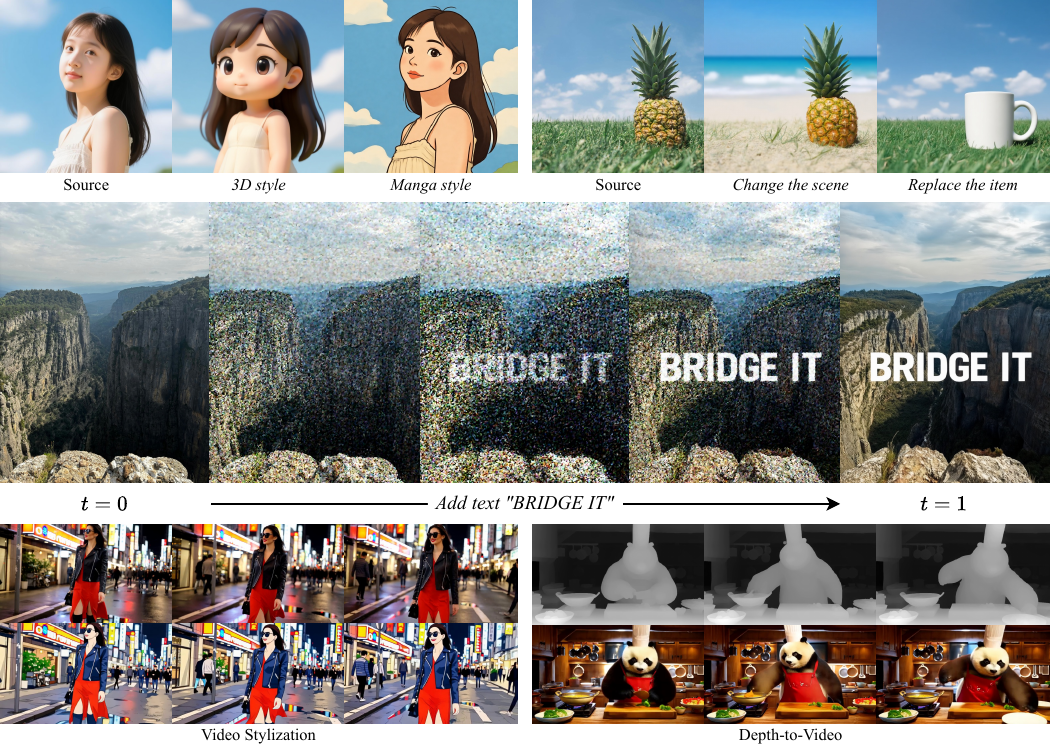}
    \vspace{-0.5cm}
    \caption{Results of vision bridge transformer on various vision translation tasks.}
    \vspace{1em}
\end{center} 
}]

{\renewcommand{\thefootnote}{\fnsymbol{footnote}} \footnotetext[1]{Corresponding author. (\texttt{xinchao@nus.edu.sg})}}

\begin{abstract}
We introduce \textbf{Vision Bridge Transformer (ViBT)}, a large-scale instantiation of Brownian Bridge Models designed for conditional generation.
Unlike traditional diffusion models that transform noise into data, Bridge Models directly model the trajectory between inputs and outputs, creating an efficient data-to-data translation paradigm.
By scaling these models to 20B and 1.3B parameters, we demonstrate their effectiveness for image and video translation tasks.
To support this scale, we adopt a Transformer architecture and propose a variance-stabilized velocity-matching objective for robust training.
Together, these advances highlight the power of scaling Bridge Models for instruction-based image editing and complex video translation.
\end{abstract}
\section{Introduction}
\label{sec:intro}

Generative models have advanced remarkably, beginning with Generative Adversarial Networks (GANs)~\cite{goodfellow2014generative, karras2020analyzing} that enabled high-quality image synthesis via adversarial training. More recently, probability-path methods, especially diffusion models~\cite{ho2020denoising, song2020score, lipman2022flow}, have further elevated generative capabilities. Transformer-based architectures trained at scale have significantly enhanced the fidelity and diversity of synthesized images~\cite{peebles2023scalable, esser2024scaling, chen2023pixart, wu2025qwen} and videos~\cite{wan2025wan, team2025longcat, chen2025sana}.
Building upon this success, extending these models to conditional vision generation tasks has become a natural direction~\cite{zhang2023adding, ye2023ip, tan2025ominicontrol, tan2025ominicontrol2}.
Typically, these approaches inject visual conditions into the generative process by incorporating source images or videos as auxiliary inputs~\cite{zhang2023adding, ye2023ip}.

Despite their success, the underlying \textit{noise-to-vision} modeling paradigm widely adopted by these models~\cite{ esser2024scaling, esser2024scaling,goodfellow2014generative} can be unnatural for conditional generation tasks.
In this paradigm, models start from noise and gradually refine it toward the target~\cite{chen2023pixart, wan2025wan, labs2025flux, wu2025qwen}.
However, in many conditional scenarios (e.g., image editing, colorization, and frame interpolation), inputs already closely resemble the desired outputs, making this process unintuitive. 
Moreover, incorporating additional conditioning tokens introduces substantial computational overhead under transformer architectures~\cite{tan2025ominicontrol, tan2025ominicontrol2}, especially in video settings~\cite{wan2025wan, aigc_apps_videox_fun_2025, jiang2025vace}.

In contrast, the \textit{vision-to-vision} paradigm provides a more intuitive alternative by directly modeling the transformation between structured source and target domains~\cite{wang2024framebridge, li2023bbdm}.
Unlike the noise-to-vision approach, it explicitly models the direct path from conditioning inputs to target outputs, naturally capturing the strong correlations inherent in the data.
Previous works demonstrated the feasibility of vision-to-vision modeling using \textit{Bridge Models}~\cite{liu20232, li2023bbdm, Zhou2023DenoisingDB}, which construct stochastic processes connecting source and target data distributions. 
Although Brownian Bridge based formulations~\cite{li2023bbdm} have shown promising results, prior work has largely been limited to small-scale architectures and relatively simple tasks~\cite{yue2023image, zheng2024diffusion, wang2024score, chadebec2025lbm}, leaving their potential for complex vision translation scenarios underexplored~.

This work introduces \textit{Vision Bridge Transformer} (ViBT), the first Brownian Bridge Model scaled to large-scale settings for complex vision translation tasks.
ViBT leverages transformer architectures~\cite{peebles2023scalable, vaswani2017attention} initialized from leading flow-matching models~\cite{wu2025qwen, wan2025wan}, inheriting strong generative priors.
By scaling ViBT to 20B and 1.3B parameters, we demonstrate the feasibility of applying Bridge Models to previously unexplored tasks within this framework.

However, scaling Bridge Models to such large-scale architectures necessitates a robust training objective. We observe that conventional displacement-style targets~\cite{li2023bbdm} disproportionately bias training toward early generation steps, while naive velocity-based objectives~\cite{chadebec2025lbm, liu2022flow} exhibit severe numerical instability, negatively affecting convergence and performance.
To address these issues, we propose  \textit{variance-stabilized velocity matching} objective, which maintains numerical stability and equally emphasizes learning across all timesteps, facilitating efficient training at scale.


Extensive experiments demonstrate that ViBT effectively generalizes to a wide range of complex vision translation tasks, including instruction-based image editing, instruction-guided video stylization, depth-to-video synthesis, image coloring, and video frame interpolation, achieving results competitive with traditional conditional diffusion methods while being significantly more efficient. 
Comprehensive ablation studies further verify the effectiveness of the variance-stabilized velocity matching objective.


\section{Related Works}
\label{sec:related_works}

\paragraph{Generative models}
Early generative models, such as Variational Autoencoders (VAEs)~\cite{pu2016variational} and Generative Adversarial Networks (GANs)~\cite{goodfellow2014generative, karras2019style}, enabled high-quality synthesis through latent modeling and adversarial training.
Diffusion models~\cite{ho2020denoising, rombach2022high} later introduced iterative denoising processes, significantly advancing image and video generation.
Flow-matching models~\cite{esser2024scaling, lipman2022flow} further reframed generation as learning deterministic or stochastic paths between distributions.
More recently, transformer-based architectures trained at scale have further enhanced the fidelity and diversity of generative models~\cite{peebles2023scalable, wu2025qwen, wan2025wan}.

\vspace{-0.5em}
\paragraph{Conditional generation}
Conditional diffusion models incorporate auxiliary signals such as text, images, poses, or depth maps through additional encoders, auxiliary branches, or cross-attention mechanisms.
Representative methods include ControlNet~\cite{zhang2023adding}, IP-Adapter~\cite{ye2023ip}, and T2I-Adapters~\cite{mou2024t2i}.
With the emergence of Diffusion Transformers (DiT)~\cite{peebles2023scalable}, recent approaches~\cite{tan2025ominicontrol, labs2025flux, wu2025qwen} incorporate conditions directly into transformer attention layers for stronger guidance.
However, these methods introduce substantial computational overhead, especially in video tasks.


\vspace{-0.5em}
\paragraph{Bridge models}
Bridge Models~\cite{liu20232, li2023bbdm, Zhou2023DenoisingDB} construct stochastic processes directly connecting source and target distributions, providing an alternative to noise-driven generation. Early approaches employed Schrödinger bridges~\cite{de2021diffusion}, stochastic interpolants~\cite{albergo2023stochastic}, and entropic optimal transport~\cite{cuturi2013sinkhorn}.
Recent diffusion-based variants, such as Denoising Diffusion Bridge Models (DDBM)~\cite{Zhou2023DenoisingDB} and Brownian Bridge methods~\cite{li2023bbdm}, demonstrated promising results for conditional generation and image translation tasks.

Several recent works have highlighted the potential of Bridge Models in vision tasks, demonstrating improved structural and stylistic consistency in exemplar-guided image translation~\cite{lee2024ebdm}, enhanced temporal coherence in video synthesis~\cite{Vasilev2025TimeCorrelatedVB}, and increased efficiency during training and inference for basic image translation tasks~\cite{Ji2024DPBridgeLD, chadebec2025lbm}.


\newpage
\section{Preliminaries}
\label{sec:preliminaries}

Probability path modeling~\cite{lipman2022flow, liu2022flow, de2021diffusion} defines a class of generative models that describe continuous-time processes transporting mass from a prior distribution \( p_0 \) to a target distribution \( p_1 \). Generally, these models are represented by a stochastic differential equation (SDE):
\begin{equation}
\label{eq:general-sde}
\mathrm{d}X_t = v(X_t, t)\,\mathrm{d}t + \sigma(t)\,\mathrm{d}W_t,\quad t\in[0,1],
\end{equation}
with boundary conditions \( X_0\sim p_0 \) and \( X_1\sim p_1 \), velocity field \( v:\mathbb{R}^d\times[0,1]\to\mathbb{R}^d \), diffusion coefficient \(\sigma:[0,1]\to\mathbb{R}_{\ge 0}\), and standard Brownian motion \( W_t \).

In practice, the velocity field is typically parameterized by a neural network \( v_\theta \), trained via a velocity-matching objective~\cite{lipman2022flow, liu2022flow}:
\begin{equation}
\label{eq:vm-loss}
\mathcal{L}(\theta)=\mathbb{E}_{(x_0,x_1),\,t,\,X_t}\bigl[\|v_\theta(X_t,t)-u_t(X_t\,|\,x_0,x_1)\|^2\bigr],
\end{equation}
where \( u_t(\cdot\,|\,x_0,x_1) \) denotes the instantaneous velocity induced by a chosen teacher trajectory (deterministic or stochastic), and \( X_t \) is sampled accordingly. Throughout, we use uppercase \( X_t \) for stochastic states and lowercase \( x_t \) for deterministic trajectories.

\paragraph{Rectified Flow}
Rectified Flow~\cite{liu2022flow, esser2024scaling} is a deterministic realization of probability path modeling obtained by setting \(\sigma(t)\equiv 0\) in Eq.~\eqref{eq:general-sde}. It defines linear deterministic trajectories connecting endpoints \(x_0\sim p_0\), typically sampled from a standard Gaussian distribution \(\mathcal{N}(0, I)\), and \(x_1\sim p_1\), drawn from the data distribution:
\begin{equation}
\label{eq:rf-path}
x_t = (1 - t)x_0 + t x_1.
\end{equation}
Under this linear interpolation, the instantaneous velocity target simplifies to a constant vector:
\begin{equation}
u_t\equiv x_1 - x_0.
\end{equation}

\paragraph{Brownian Bridge}
In contrast to the deterministic Rectified Flow, the standard Brownian Bridge~\cite{albergo2023stochastic, li2023bbdm} incorporates stochasticity via a constant diffusion coefficient \(\sigma(t)\equiv1\). Given fixed endpoints \( (x_0,x_1) \), its conditional intermediate states follow a Gaussian distribution:
\begin{equation}
\label{eq:bb-conditional}
X_t \mid_{(x_0,x_1)} 
\sim 
\mathcal{N}\!\bigl(
\underbrace{(1 - t)x_0 + t x_1}_{\text{linear interpolation}},\;
\underbrace{t(1 - t) I}_{\text{maximal variance at }t=0.5}
\bigr).
\end{equation}
Brownian Bridges are particularly suited to data-to-data transport tasks, such as denoising corrupted samples or translating data between structured image and video domains. Pairs of endpoints \( (x_0,x_1) \) are sampled from their respective source and target distributions. Under this stochastic formulation, the instantaneous velocity target used in velocity matching is expressed as:
\begin{equation}
u_t(X_t\,|\,x_0,x_1)=\frac{x_1 - X_t}{1 - t}.
\label{eq:velocity-bb}
\end{equation}
\section{Methodology}
\label{sec:methodology}
Our method leverages a transformer-based architecture to model vision translation tasks in latent space.
Given paired source and target data (images or videos), we encode them into latent representations $x_0 \sim p_\text{source}$ and $x_1 \sim p_\text{target}$ using a pre-trained VAE encoder~\cite{kingma2013auto}, and apply the Brownian Bridge formulation to directly model the transformation from $x_0$ to $x_1$.

\subsection{Stabilized velocity matching}
\begin{figure}[t]
    \centering
    \includegraphics[width=1\columnwidth]{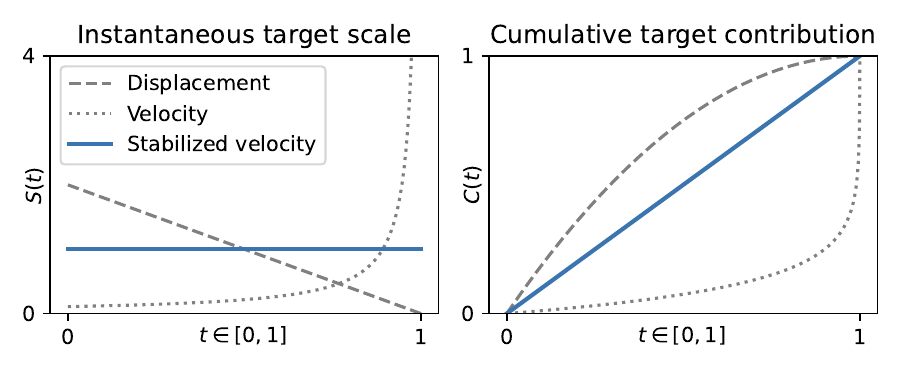}
    \caption{\textbf{Instantaneous and cumulative target contributions.}
$S(t)=\mathbb{E}\|\tau_t\|^2$ with $\tau_t\in\{d_t,u_t,\tilde u_t\}$.
$C(t)=\frac{\int_0^{t}S(s)\,ds}{\int_0^{0.999}S(s)\,ds}$.
}
    \label{fig:loss_contributions}
\end{figure}

During training, given latent endpoint pairs $(x_0,x_1)\sim p_\text{source,target}$, we uniformly sample a time $t\in[0,1]$ and Gaussian noise $\epsilon\sim\mathcal{N}(0,I)$.
According to the conditional distribution of the Brownian bridge~\eqref{eq:bb-conditional}, the intermediate latent state $x_t$ is constructed as:
\begin{equation}
    x_t = (1 - t)x_0 + t x_1 + \sqrt{t(1 - t)}\,\epsilon.
\end{equation}
The velocity-based training target at this noisy state, derived from Eq.~\eqref{eq:velocity-bb}, is given by:
\begin{equation}
    u_t(x_t|x_1) = \frac{x_1 - x_t}{1 - t} = (x_1 - x_0) - \sqrt{\frac{t}{1 - t}}\,\epsilon.
\end{equation}
Accordingly, the training objective is given by the mean squared error between the predicted and target velocities:
\begin{equation}
    \mathcal{L}_{\text{velocity}} = 
    \mathbb{E}_{t,\,\epsilon,\,x_0,\,x_1}
    \left[
        \left\|v_\theta(x_t, t) - u_t(x_t|x_1)\right\|^2
    \right].
\end{equation}
However, this objective faces critical issues as $t\to 1$: the target velocity $u_t(x_t|x_1)$ diverges as $O\left(\frac{1}{\sqrt{1 - t}}\right)$, causing instability, and the loss excessively focuses on these states, neglecting intermediate ones (Fig.~\ref{fig:loss_contributions}).

An alternative approach adopted in previous works~\cite{li2023bbdm} is to use a displacement-based training target defined as
\begin{equation}
    d_t(x_t|x_1) = x_1 - x_t.
\end{equation}
Accordingly, the displacement-based training objective is given by the mean squared error:
\begin{equation}
    \label{eq:displacement-loss}
    \mathcal{L}_{\text{displacement}} = \mathbb{E}_{t,\,\epsilon,\,x_0,\,x_1}\left[\|d_\theta(x_t, t) - d_t(x_t|x_1)\|^2\right].
\end{equation}
This displacement formulation naturally avoids numerical divergence, as it remains stable across all timesteps. However, its magnitude diminishes as $t\to1$ at the rate $O(\sqrt{1-t})$, causing the training loss to be dominated by samples at smaller values of $t$.

Motivated by the above numerical instability and imbalanced loss across timesteps, we propose \textit{stabilized velocity matching}, which introduces a normalization factor $\alpha(x_0,x_1,t)$ to balance loss contributions across timesteps. 
We rescale the original velocity target as:
\begin{equation}
    \tilde{u}_t(x_t|x_1) = \frac{u_t(x_t|x_1)}{\alpha(x_0,x_1,t)}.
\end{equation}
Specifically, we define $\alpha(x_0,x_1,t)$ based on the normalized root mean square magnitude of the velocity:
\begin{align}
    \alpha(x_0,x_1,t)^2 
    &= \frac{\mathbb{E}\left[\|u_t(x_t|x_1)\|^2\right]}{\|x_1 - x_0\|^2} \\
    &= 1 + \frac{t\,D}{(1 - t)\,\|x_1 - x_0\|^2},
\end{align}
where $D$ denotes the latent dimensionality
\footnote{Derivations of factor $\alpha(\cdot)$ are in the Supplementary Material~\ref{sec:theoretical-derivations}.}
. As shown in Fig.~\ref{fig:loss_contributions}, this choice significantly reduces divergence and ensures balanced loss contributions throughout training.

The resulting stabilized velocity-matching objective is:
\begin{align}
    \label{eq:stabilized-velocity-loss}
    \mathcal{L}_{\tilde{\text{velocity}}} 
    &= \mathbb{E}_{t,\,\epsilon,\,x_0,\,x_1}\left[
        \bigl\|\tilde{v}_\theta(x_t, t)-\tilde{u}_t(x_t|x_1)\bigr\|^2
    \right],
\end{align}
where $\tilde{v}_\theta(x_t, t)=v_\theta(x_t, t)/\alpha(x_0,x_1,t)$ normalizes the network prediction for loss calculation only, while the network still directly predicts velocity.
The complete training procedure is summarized in Algorithm~\ref{alg:training}.

\begin{algorithm}[t!]
    \caption{Training}
    \label{alg:training}
    \KwIn{data pairs $(x_0,x_1)\sim p_{\text{source,target}}$, model $v_\theta$, latent dimension $D$}
    \Repeat{convergence}{
        Sample latent pair $(x_0,x_1)$, interpolation time $t\sim U(0,1)$, and noise $\epsilon\sim\mathcal{N}(0,I)$\;
        Construct intermediate state $x_t = (1-t)x_0 + t x_1 + \sqrt{t(1-t)}\,\epsilon$\;
        Compute velocity target $u_t = (x_1 - x_t)/(1-t)$\;
        Compute normalization factor $\alpha^2 = 1 + {tD}/{[(1 - t)\|x_1 - x_0\|^2]}$\;
        Compute stabilized velocity loss $\mathcal{L}_{\tilde{\text{velocity}}} = \|\frac{v_\theta(x_t,t)-u_t}{\alpha}\|^2$\;
        Update model parameters $\theta$ by gradient descent on $\mathcal{L}_{\tilde{\text{velocity}}}$\;
    }
\end{algorithm}

\newpage
\subsection{Variance-corrected sampling}
\begin{algorithm}[t!]
    \caption{Inference}
    \label{alg:inference}
    \KwIn{source-target latent pair $(x_0,x_1)$, trained model $v_\theta$, latent dimension $D$, discretization steps $N$, discretization schedule $0=t_0<t_1<\dots<t_N=1$}
    Initialize $x \gets x_0$\;
    \For{$k=0,1,\dots,N-1$}{
        Compute step size $\Delta t \gets t_{k+1}-t_k$\;
        Compute scaling factor $\eta \gets \sqrt{\Delta t \frac{1 - t_{k+1}}{1 - t_k}}$\;
        Sample noise $\epsilon\sim\mathcal{N}(0,I)$\;
        Update latent state:
        \[x \gets x + \Delta t\,v_\theta(x, t_k) + \eta\,\epsilon\]
    }
    \KwOut{Final state $x$ approximating the target $x_1$}
\end{algorithm}

To sample from the trained Brownian Bridge model with stabilized velocity matching, we discretize the continuous-time SDE defined in Eq.~\eqref{eq:general-sde} using the Euler-Maruyama discretization~\cite{maruyama1955continuous}.
Given a schedule $0 = t_0 < t_1 < \dots < t_N = 1$, the sampling process starts from the source $x_0$ and iteratively updates the latent state towards the target $x_1$.

The standard Euler--Maruyama discretization yields:
\begin{equation}
    x_{k+1}^{\text{standard}}
    = x_k + \Delta t_k\, v_\theta(x_k, t_k)
    + \sqrt{\Delta t_k}\,\epsilon_k,
\end{equation}
where $\Delta t_k = t_{k+1}-t_k$ and 
$\{\epsilon_k\}_{k=0}^{K-1}$ are i.i.d.\ samples drawn from $\mathcal{N}(0, I)$.
This scheme assumes a locally constant variance structure, i.e., the stochastic term scales purely with $\sqrt{\Delta t_k}$.

However, in the Brownian Bridge process, the variance should gradually shrink as the trajectory approaches the target $x_1$, reflecting decreasing uncertainty near $t=1$.
Consequently, the noise magnitude in the naive discretization becomes overly large at late steps, leading to biased trajectories and degraded sample quality.

To correct this mismatch, a scaling factor can be applied to continuously modulate the variance across timesteps.
In practice, the diffusion term is rescaled by the ratio $\frac{1 - t_{k+1}}{1 - t_k}$\footnote{Derivations of the scaling ratio are in the Supplementary Material~\ref{sec:theoretical-derivations}.}, resulting in a variance-corrected stochastic update~\cite{albergo2023stochastic, li2023bbdm}:
\begin{equation}
    \label{eq:stoch-update}
    x_{k+1}^{\text{corrected}} = x_k + \underbrace{\Delta t_k\,v_\theta(x_k, t_k)}_{\text{velocity toward target}} + \underbrace{\sqrt{\Delta t_k\frac{1 - t_{k+1}}{1 - t_k}}\,\epsilon_k}_{\text{variance-corrected noise}}.
\end{equation}
This correction ensures that the variance decays smoothly as $t\!\to\!1$, aligning the discrete sampling dynamics with the intrinsic structure of the Brownian Bridge.
The complete inference procedure is summarized in Algorithm~\ref{alg:inference}.

\section{Experiments}
\label{sec:experiments}

We conduct extensive experiments to explore the effectiveness of scaling Brownian Bridge diffusion models across various complex vision conditional generation tasks.
We begin with instruction-based image editing tasks in Section~\ref{sec:image-editing} to evaluate the model's ability to perform fine-grained and instruction-based content modification.
Next, we extend our study to video stylization in Section~\ref{sec:video-stylization}, where input videos are transformed into target styles specified by textual instructions while preserving the original motion and structure.
Finally, we examine video translation tasks focusing primarily on depth-to-video synthesis in Section~\ref{sec:experiments_translation}.
Additionally, detailed ablation studies in Section~\ref{sec:experiments_ablation} validate the effectiveness of our proposed stabilized velocity-matching loss and explore key properties of the scaled Brownian Bridge diffusion process.

\paragraph{Training and inference details}
For image and video modalities, we respectively initialize our models from state-of-the-art pre-trained models: Qwen-Image-Editing~\cite{wu2025qwen} with 20B parameters for image-based tasks, and Wan 2.1~\cite{wan2025wan} with 1.3B parameters for video-based tasks. 
During training, the image model employs LoRA~\cite{hu2022lora} with a rank of 128, while the video model undergoes full-parameter updates.
We train our models using the Prodigy optimizer~\cite{mishchenko2024prodigy} with a learning rate of 1 and set \texttt{save\_warmup=True}. 
By default, we train each model for 20,000 iterations on 1 NVIDIA H100 GPU with a batch size of 1.

\subsection{Instruction-based image editing}
\label{sec:image-editing}
We first evaluate our bridge models on the complex image editing task, which involves modifying specific content within an input image based on textual instructions while preserving other regions.
In this task, the input image serves as the source domain $p_{\rm source}$, and the edited image represents the target domain $p_{\rm target}$. The brownian bridge model directly learns the transformation between the base image and the edited output.

\paragraph{Dataset}
We create a synthetic dataset for instruction-based image editing based on the \textit{Open Images Dataset}~\cite{kuznetsova2020open}.
Specifically, we first randomly sample 5,000 images and generate corresponding editing instructions using the vision-language model Qwen3-VL~\cite{yang2025qwen3}.
We then produce edited images based on these instructions using the Qwen-Image-Editing model~\cite{wu2025qwen}.
Additionally, we enrich our dataset by incorporating stylized image data generated by OmniConsistency~\cite{song2025omniconsistency}.
Finally, we filter the generated instruction-image pairs with Qwen3-VL to ensure high alignment between the instructions and the image edits. The detailed dataset construction process is described in the Supplementary Material, Section~\ref{sec:exp_detail}.

\paragraph{Evaluation and baselines} 
We adopt the \textit{ImgEdit-Bench}~\cite{ye2025imgedit} as our evaluation benchmark, as it provides a comprehensive assessment across multiple editing dimensions, including instruction-following accuracy, editing quality, and preservation of image details. All evaluations presented in this section strictly follow the official protocols defined by ImgEdit-Bench. 
We compare our bridge model with representative diffusion-based methods, including InstructPix2Pix~\cite{brooks2023instructpix2pix}, Qwen-Image-Editing~\cite{wu2025qwen}, Step1X-edit~\cite{liu2025step1x}, FLUX.1 Kontext~\cite{labs2025flux}, and several other notable approaches~\cite{zhang2023magicbrush, yu2025anyedit, li2025uniworld}.

\begin{figure}[t!]
    \centering
    \includegraphics[width=\linewidth]{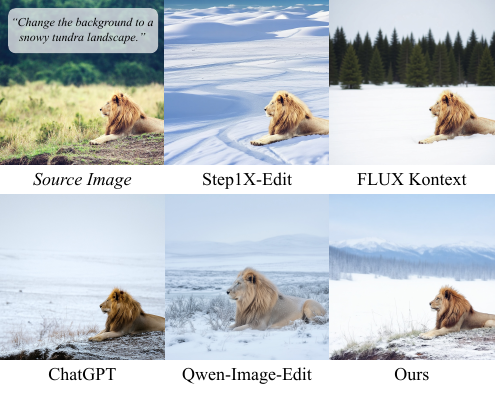}
    \caption{Qualitative comparison on image editing.}
    \label{fig:image-comparison}
\end{figure}

\begin{table}[t]
\centering
\setlength{\tabcolsep}{3pt}  
\resizebox{\columnwidth}{!}{
\begin{tabular}{l|ccccccccc|c}
\toprule
\textbf{Model} & Add & Adjust & Extract & Replace & Remove & Bg. & Style & Hybrid & Action & \textbf{Avg.} \\
\midrule
MagicBrush & 2.84 & 1.58 & 1.51 & 1.97 & 1.58 & 1.75 & 2.38 & 1.62 & 1.22 & 1.83 \\
Ins.Pix2Pix & 2.45 & 1.83 & 1.44 & 2.01 & 1.50 & 1.44 & 3.55 & 1.20 & 1.46 & 1.88 \\
AnyEdit & 3.18 & 2.95 & 1.88 & 2.47 & 2.23 & 2.24 & 2.85 & 1.56 & 2.65 & 2.45 \\
Step1X-Edit & 3.88 & 3.14 & 1.76 & 3.40 & 2.41 & 3.16 & 4.63 & 2.64 & 2.52 & 3.06 \\
UniWorld-V1 & 3.82 & 3.64 & 2.27 & 3.47 & \underline{3.24} & 2.99 & 4.21 & 2.96 & 2.74 & 3.26 \\
\rowcolor{gray!20}
ViBT & \textbf{4.20} & 3.70 & 2.31 & \underline{3.86} & 2.91 & 3.92 & \underline{4.85} & 2.72 & 3.52 & 3.55 \\
FLUX Kontext & 3.82 & 3.64 & 2.27 & 3.47 & \underline{3.24} & 2.99 & 4.21 & 2.96 & 2.74 & 3.71 \\
\rowcolor{gray!20}
ViBT ($s=0.5$)\tablefootnote{The noise scale is set to $s=0.5$; details are given in Section~\ref{sec:experiments_ablation}.}& 4.14 & \underline{4.20} & \textbf{2.64} & 3.72 & 3.03 & \underline{4.06} & \textbf{4.87} & \underline{3.19} & \underline{3.95} & \underline{3.76} \\
Qwen-Image-Edit & \underline{4.17} & \textbf{4.29} & \underline{2.44} & \textbf{4.30} & \textbf{3.90} & \textbf{4.15} & 4.00 & \textbf{3.32} & \textbf{4.51} & \textbf{3.90} \\
\bottomrule
\end{tabular}}
\caption{Model ranking on \textit{ImgEdit-Bench} based on average score.}
\label{tab:imgedit_benchmark}
\end{table}

\begin{figure*}[t!]
    \centering
    \includegraphics[width=\linewidth]{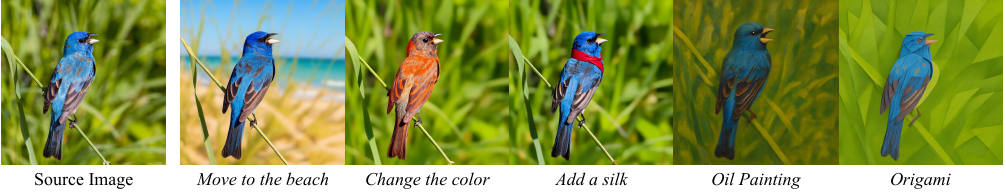}
    \caption{Qualitative results of the image editing.}
    \label{fig:image-editing}
\end{figure*}

\paragraph{Results and analysis} 
Table~\ref{tab:imgedit_benchmark} reports the quantitative results on ImgEdit-Bench\footnote{Results of baselines are reported from ImgEdit Bench~\cite{ye2025imgedit, lin2025uniworld, li2025uniworld}.}. 
ViBT performs on a similar level to current state-of-the-art methods across the different editing categories.
In tasks such as object addition and style transfer, ViBT achieves notably stronger results, outperforming competing approaches by a clear margin. 
The qualitative results in Figure~\ref{fig:image-comparison} and \ref{fig:image-editing} show that ViBT produces clear instruction-following edits while keeping the original scene content, achieving visual quality comparable to leading diffusion-based methods.

\subsection{Video stylization}
\label{sec:video-stylization}

\begin{figure}[t]
    \centering
    \includegraphics[width=\columnwidth]{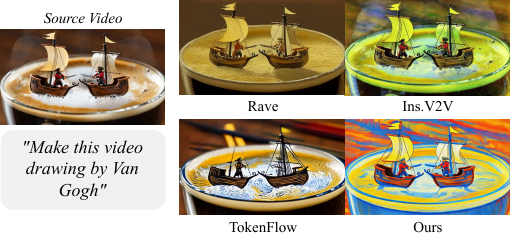}
    \caption{Comparison of stylized videos under the \textit{Van Gogh} style.}
    \label{fig:video-stylization-compare}
\end{figure}
In the video domain, we first consider the instruction-based video stylization task, which aims to modify the visual style of an input video according to a given textual instruction while preserving its original content and motion dynamics.

\begin{table}[t]
\centering
\setlength{\tabcolsep}{3pt}
\resizebox{\columnwidth}{!}{
\begin{tabular}{lcccccc}
\toprule
\textbf{Method} & NIQE $\downarrow$ & TOPIQ$_{\rm NR}$ $\uparrow$ & MUSIQ $\uparrow$ & MANIQA $\uparrow$ & CLIPIQA $\uparrow$ & \makecell{CLIP\\Score} $\uparrow$\\
\midrule
TokenFlow & 4.767 & 0.376 & 55.186 & 0.267 & 0.378 & 0.683 \\
Ins.V2V   & \textbf{4.268} & \underline{0.467} & \underline{60.621} & \underline{0.306} & \underline{0.440} & \textbf{0.827} \\
RAVE      & 6.514 & 0.351 & 50.595 & 0.269 & 0.377 & 0.683 \\
\rowcolor{gray!20}
ViBT      & \underline{4.328} & \textbf{0.503} & \textbf{64.045} & \textbf{0.348} & \textbf{0.486} & \underline{0.782} \\
\bottomrule
\end{tabular}}
\caption{Quantitative results on the video stylization task.}
\label{tab:video-stylization}
\end{table}

\paragraph{Dataset and training}
We use the open-source \textit{Ditto-1M} dataset~\cite{bai2025ditto} for training our bridge video model. Specifically, we randomly sample 10,000 video samples from the subset \textit{global\_style1} of Ditto-1M, which contains videos paired with style descriptions. We train bridge model on 4 NVIDIA H100 GPUs for 50,000 iterations in this task.

\paragraph{Evaluation and baselines}
For evaluation, we construct a benchmark comprising 100 videos generated by \textit{Wan 2.2 14B}~\cite{wan2025wan} using the first 100 prompts from \textit{MovieGen Bench}~\cite{polyak2024movie}, serving as inputs for the stylization task.
These videos do not overlap with our training set.
Each video is paired with a randomly sampled textual style instruction.
We stylize videos consisting of 81 frames each, and uniformly sample 5 frames per video for quality assessment.
We quantitatively evaluate these sampled frames using widely adopted image-quality metrics, including NIQE~\cite{mittal2012making}, TOPIQ-NR~\cite{chen2024topiq}, MUSIQ~\cite{ke2021musiq}, MANIQA~\cite{yang2022maniqa}, CLIPIQA~\cite{wang2023exploring}, and CLIPScore~\cite{hessel2021clipscore}.
These metrics comprehensively measure perceptual image quality, aesthetic appeal, and visual-semantic alignment with textual instructions.
We compare our method against three diffusion-based video stylization methods, \textit{Instruct Video-to-Video (InsV2V)}~\cite{cheng2023consistent}, \textit{RAVE}~\cite{kara2024rave}, and \textit{TokenFlow}~\cite{qu2024tokenflow}.

\begin{figure}[t]
    \centering
    \includegraphics[width=\columnwidth]{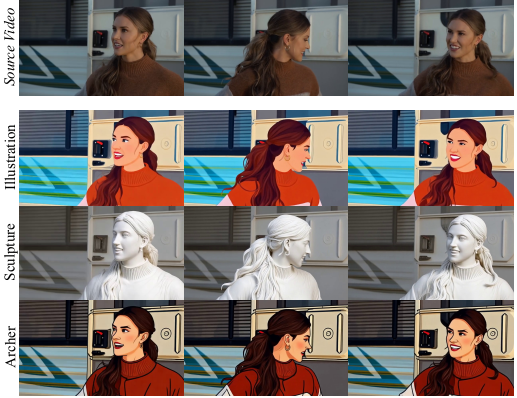}
    \caption{Qualitative comparisons of ViBT on the video stylization task with different styles.}
    \label{fig:video-stylization}
\end{figure}

\paragraph{Results and analysis}
Quantitative results in {Table~\ref{tab:video-stylization}} show that ViBT outperforms the baselines in most metrics, demonstrating its effectiveness in generating high-quality stylized videos that align well with the given instructions.
Qualitative comparisons in Figure~\ref{fig:video-stylization-compare} further illustrate that ViBT can apply the desired style to the input video while preserving the original motion and structure.
Figure~\ref{fig:video-stylization} futher demonstrates that ViBT can effectively stylize videos across various artistic styles while preserving original content and motion.
More stylized video examples are available in the Supplementary Material, Section~\ref{sec:additional_results}.

\begin{table*}[t]
    \centering
    \scriptsize
    \renewcommand{\arraystretch}{1.15}
    \arrayrulecolor{black}
    \begin{adjustbox}{width=\textwidth}
    \begin{tabular}{l|c|ccccc|ccc|c|c}
    \toprule
    \multirow[c]{2}{*}{\textbf{Method}} &
    \multirow[c]{2}{*}{\textbf{Base Model}} &
    \multicolumn{5}{c|}{\textbf{Perceptual quality}} &
    \multicolumn{3}{c|}{\textbf{Ground truth similarity}} &
    \multirow[c]{2}{*}{\textbf{CLIP Score}$\uparrow$} &
    \multirow[c]{2}{*}{\textbf{VBench Score}$\uparrow$} \\
    \cmidrule(lr){3-7} \cmidrule(lr){8-10}
     & &
     NIQE$\downarrow$ &
     TOPIQ$_{\rm NR}$$\uparrow$ &
     MUSIQ$\uparrow$ &
     MANIQA$\uparrow$ &
     CLIPIQA$\uparrow$ &
     SSIM$_{\rm c}$$\uparrow$ &
     PSNR$\uparrow$ &
     DISTS$\downarrow$ &
     & \\
    \midrule
    ControlVideo & SD 1.5
        & 6.641
        & {0.443}
        & {50.735}
        & \textbf{0.354}
        & {0.436}
        & 0.385
        & 9.067
        & 0.465
        & 0.732
        & 0.62 \\
    Control A Video & SD 1.5
        & \underline{5.102}
        & 0.374
        & 52.391
        & 0.254
        & 0.334
        & 0.276
        & 8.510
        & 0.348
        & 0.715
        & 0.59 \\
    VideoComposer & SD 2.1
        & 6.750
        & 0.305
        & 43.691
        & 0.276
        & 0.237
        & 0.329
        & 9.656
        & 0.457
        & 0.722
        & 0.59 \\
    Wan Fun Control & Wan 2.1
        & 5.346
        & \textbf{0.477}
        & \underline{59.086}
        & \underline{0.335}
        & \underline{0.459}
        & \underline{0.427}
        & \underline{10.899}
        & \underline{0.281}
        & \underline{0.776}
        & \underline{0.69} \\
    \rowcolor{gray!20}
    ViBT & Wan 2.1
        & \textbf{4.896}
        & \textbf{0.477}
        & \textbf{59.625}
        & 0.331
        & \textbf{0.477}
        & \textbf{0.429}
        & \textbf{11.403}
        & \textbf{0.230}
        & \textbf{0.781}
        & \textbf{0.71} \\
    \bottomrule
    \end{tabular}
    \end{adjustbox}
    \caption{Quantitative comparison on the \textit{depth-to-video} task.}
    \label{tab:quantitative2}
\end{table*}

\begin{table*}[t]
\centering
\scriptsize
\setlength{\tabcolsep}{3.5pt}
\renewcommand{\arraystretch}{1.15}
\arrayrulecolor{black}
\begin{adjustbox}{width=\linewidth}
\begin{tabular}{l|cccccccccccccccc|c}
\toprule
\textbf{Method} &
\makecell{Subj.\\Cons.} &
\makecell{Bkgd.\\Cons.} &
\makecell{Aesth.\\Qual.} &
\makecell{Img.\\Qual.} &
\makecell{Obj.\\Class} &
\makecell{Multi\\Objs.} &
\makecell{Color} &
\makecell{Spatial\\Rel.} &
\makecell{Scene} &
\makecell{Temp.\\Style} &
\makecell{Overall\\Cons.} &
\makecell{Human\\Action} &
\makecell{Temp.\\Flicker} &
\makecell{Motion\\Smooth.} &
\makecell{Dyn.\\Degree} &
\makecell{Appear.\\Style} &
\makecell{\textbf{Avg.}\\\textbf{Score}} \\
\midrule
Control Video
  & 0.899 & \textbf{0.94} & 0.54 & 0.52 & 0.57 & 0.26 & 0.706 & 0.46 & 0.29 & 0.20 & {0.24} & 0.80 & 0.991 & \textbf{0.990} & 0.11 & \textbf{0.229} & 0.55 \\
Control A Video
  & 0.791 & 0.88 & 0.48 & \underline{0.59} & 0.59 & 0.25 & 0.799 & 0.44 & 0.43 & 0.21 & 0.24 & 0.83 & 0.982 & 0.976 & 0.72 & \underline{0.235} & 0.59 \\
Video Composer
  & 0.873 & 0.92 & 0.44 & 0.48 & 0.67 & 0.23 & \textbf{0.854} & 0.32 & 0.29 & 0.22 & 0.24 & 0.91 & 0.963 & 0.949 & \textbf{0.88} & 0.222 & 0.59 \\
Wan Fun
  & \textbf{0.913} & 0.93 & \underline{0.60} & 0.57 & \underline{0.87} & \underline{0.65} & \underline{0.848} & \underline{0.70} & \underline{0.46} & \underline{0.24} & \underline{0.26} & \textbf{1.00} & \underline{0.989} & \underline{0.978} & \underline{0.86} & 0.211 & \underline{0.69} \\
\rowcolor{gray!20}
{ViBT}
  & \underline{0.907} & \underline{0.93} & \textbf{0.63} & \textbf{0.63} & \textbf{0.91} & \textbf{0.71} & 0.835 & \textbf{0.74} & \textbf{0.54} & \textbf{0.25} & \textbf{0.27} & \textbf{1.00} & \textbf{0.990} & 0.976 & 0.82 & 0.221 & \textbf{0.71} \\
\bottomrule
\end{tabular}
\end{adjustbox}
\caption{Quantitative comparison on the VBench attribute breakdown for the \textit{depth-to-video} task.}
\label{tab:vbench_attributes}
\end{table*}

\subsection{Video translation}
\label{sec:experiments_translation}

To verify the versatility and generalization capability of bridge model, we further explore its application to video translation tasks. We primarily investigate depth-to-video synthesis, a fundamental yet challenging scenario.

\paragraph{Dataset and training}
To create the training dataset, we first generate 1,003 videos using Wan 2.2 14B with prompts sourced from the \textit{MovieGen Bench}~\cite{polyak2024movie}.
We then transform these synthesized videos into depth maps using the \textit{Depth Anything V2}~\cite{depth_anything_v2} model, forming depth-video pairs for training.
Detailed generation procedures are provided in Supplementary Material, Section~\ref{sec:exp_detail}.

\paragraph{Evaluation and baselines}
We evaluate the brownian bridge model on the depth-to-video synthesis task, broadly adhering to the evaluation protocols outlined in \textit{VBench}~\cite{huang2023vbench}. Specifically, we first generated 946 reference videos using Wan 2.2 14B based on the prompts provided by VBench, and subsequently converted these videos into corresponding depth maps. These depth maps were employed as conditioning inputs across all methods. Further details regarding the generation procedure are provided in the Supplementary Material, Section~\ref{sec:exp_detail}. 

For comprehensive assessment, we initially applied the quality metrics discussed in Section~\ref{sec:video-stylization}. We then augmented this analysis by introducing reference-based metrics including SSIM~\cite{wang2004image}, PSNR~\cite{pyiqa}, and DISTS~\cite{ding2020image}, to quantitatively measure similarity between generated outputs and ground-truth videos. Additionally, we included the VBench Score~\cite{huang2023vbench} as an extra criterion to capture finer-grained and interpretable dimensions of video quality.

We compare ViBT against three representative diffusion-based controllable video generation models: Control-A-Video~\cite{chen2023controlavideo}, ControlVideo~\cite{zhang2023controlvideo}, and VideoComposer~\cite{wang2023videocomposer}. To provide a direct baseline for evaluating the effectiveness of our proposed method, we also incorporate Wan-Fun Control~\cite{aigc_apps_videox_fun_2025}, a flow-matching-based method initialized from the same Wan 2.1 1.3B model as ViBT.

\begin{figure}[t]
    \centering
    \includegraphics[width=\linewidth]{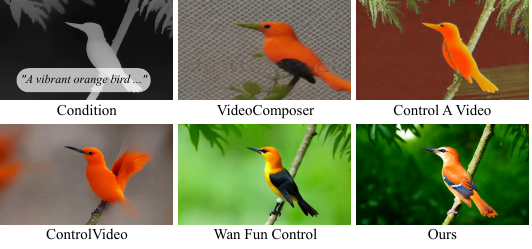}
    \caption{Qualitative comparison on the \textit{depth-to-video} task.}
    \label{fig:visualization-depth}
\end{figure}

\paragraph{Results}
Table~\ref{tab:quantitative2} presents quantitative comparisons on video frame quality, condition-following accuracy, text-following accuracy, and the overall VBench Score. 
Across all metrics, ViBT consistently outperforms the baselines, indicating strong video generation quality and reliable conditioning behavior. 
To further examine specific aspects, Table~\ref{tab:vbench_attributes} reports fine-grained attribute evaluations under VBench, where ViBT achieves leading performance on most attributes. 
Figure~\ref{fig:visualization-depth} provides qualitative examples, showing that ViBT produces richer and more detailed visuals that align more closely with the depth conditions.

Additional experimental results, including qualitative evaluations on extra video translation tasks such as video interpolation and video colorization, can be found in the Supplementary Material, Section~\ref{sec:additional_results}.
\begin{table*}[t]
\centering
\scriptsize
\arrayrulecolor{black}
\begin{adjustbox}{width=\textwidth}
\setlength{\tabcolsep}{4pt}  
\begin{tabular}{l|cccccc|cccccccccc}
\toprule
& \multicolumn{6}{c|}{\textbf{Depth-to-Video}} & \multicolumn{10}{c}{\textbf{Image Edit}} \\
\cmidrule(lr){2-7}\cmidrule(lr){8-17}
\textbf{Training Objective} & SSIM$\uparrow$ & PSNR$\uparrow$ & NIQE$\downarrow$ & DISTS$\downarrow$ & \makecell{CLIP Score$\uparrow$} & \makecell{VBench Score$\uparrow$} & Add & Adjust & Extract & Replace & Remove & Bg. & Style & Compose & Action & \multicolumn{1}{|c}{\textbf{Avg.}} \\
\midrule
Displacement
& \underline{0.409} & \underline{11.04} & \underline{4.91} & \underline{0.26} & {0.772} & {0.695}
& \underline{4.18} & \underline{3.79} & \underline{2.23} & \underline{3.57} & \underline{2.65} & \textbf{3.97} & 4.847 & \textbf{2.74} & \textbf{3.519}
& \multicolumn{1}{|c}{\underline{3.50}} \\
Velocity
& 0.428 & 10.81 & 5.45 & 0.27 & \underline{0.773} & \underline{0.698}
& 4.09 & \textbf{3.89} & 2.19 & 3.34 & 2.13 & 3.90 & \textbf{4.897} & 2.62 & 3.149
& \multicolumn{1}{|c}{3.36} \\
\rowcolor{gray!20}
Stabilized velocity
& \textbf{0.429} & \textbf{11.40} & \textbf{4.90} & \textbf{0.23} & \textbf{0.78} & \textbf{0.71}
& \textbf{4.20} & 3.70 & \textbf{2.31} & \textbf{3.86} & \textbf{2.91} & \underline{3.92} & \underline{4.850} & \underline{2.72} & \underline{3.518}
& \multicolumn{1}{|c}{\textbf{3.55}} \\
\bottomrule
\end{tabular}
\end{adjustbox}
\caption{Quantitative comparison of different training objectives.}
\label{tab:ablation-objective}
\end{table*}

\begin{table*}[t]
\centering
\scriptsize
\arrayrulecolor{black}
\begin{adjustbox}{width=\textwidth}
\setlength{\tabcolsep}{4pt}  
\begin{tabular}{l|cccccc|cccccccccc}
\toprule
& \multicolumn{6}{c|}{\textbf{Depth-to-Video}} & \multicolumn{10}{c}{\textbf{Image Edit}} \\
\cmidrule(lr){2-7}\cmidrule(lr){8-17}
\textbf{Noise Scale ($s$)} & SSIM$\uparrow$ & PSNR$\uparrow$ & NIQE$\downarrow$ & DISTS$\downarrow$ & \makecell{CLIP Score$\uparrow$} & \makecell{VBench Score$\uparrow$} & Add & Adjust & Extract & Replace & Remove & Bg. & Style & Compose & Action & \multicolumn{1}{|c}{\textbf{Avg.}} \\
\midrule
$s=0$
& 0.347 & 9.808 & 5.432 & 0.3103 & 0.717 & 0.604
& 3.91 & \textbf{4.29} & 2.01 & 2.45 & 1.60 & 3.35 & 4.65 & 2.56 & 3.07
& \multicolumn{1}{|c}{3.10} \\
$s=0.1$
& 0.331 & 9.206 & 5.413 & 0.3452 & 0.675 & 0.536
& 3.43 & 4.00 & 2.04 & 2.31 & 1.61 & 3.53 & 4.46 & 2.58 & 3.29
& \multicolumn{1}{|c}{3.03} \\
$s=0.5$
& \underline{0.398} & 10.227 & 5.185 & {0.2617} & 0.752 & {0.666}
& \underline{4.15} & \underline{4.20} & \textbf{2.64} & \underline{3.72} & \underline{3.03} & \textbf{4.06} & \textbf{4.87} & \textbf{3.19} & \textbf{3.95}
& \multicolumn{1}{|c}{\textbf{3.76}} \\
\rowcolor{gray!20}
$s=1$ (default)
& \textbf{0.429} & \textbf{11.403} & \underline{4.896} & \underline{0.2304} & \underline{0.781} & \underline{0.709}
& \textbf{4.20} & 3.70 & {2.31} & 3.86 & 2.91 & \underline{3.92} & \underline{4.85} & \underline{2.72} & \underline{3.52}
& \multicolumn{1}{|c}{\underline{3.55}} \\
$s=2$
& 0.396 & \underline{11.305} & \textbf{4.499} & \textbf{0.2295} & \textbf{0.784} & \textbf{0.711}
& 4.14 & 3.49 & \underline{2.36} & \textbf{3.94} & \textbf{3.16} & 3.64 & 4.82 & 2.46 & 2.98
& \multicolumn{1}{|c}{3.44} \\
$s=4$
& 0.394 & 10.146 & 5.912 & 0.3820 & 0.670 & 0.482
& 3.70 & 2.67 & 2.24 & 3.60 & 2.88 & 2.93 & 4.43 & 1.78 & 2.50
& \multicolumn{1}{|c}{2.97} \\
\bottomrule
\end{tabular}
\end{adjustbox}
\caption{Quantitative comparison across different noise scales ($s$).}
\label{tab:noise-scale}
\end{table*}

\begin{figure}[t]
    \centering
    \begin{subfigure}[b]{0.4\linewidth}
        \centering
        \includegraphics[width=\linewidth]{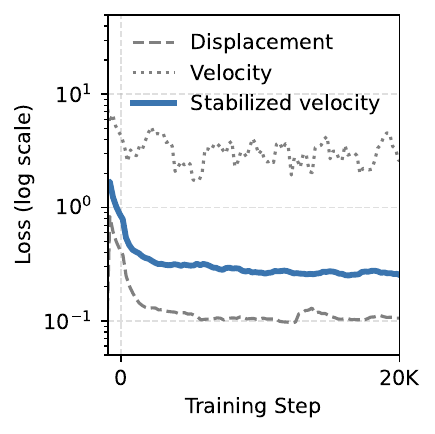}
        \caption{Training loss curves.}
        \label{fig:loss-curves}
    \end{subfigure}
    \hfill
    \begin{subfigure}[b]{0.58\linewidth}
        \centering
        \includegraphics[width=\linewidth]{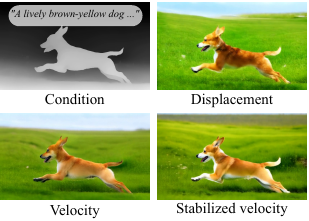}
        \caption{Visualization results.}
        \label{fig:visualization-results}
    \end{subfigure}
    \caption{Comparison of different training objectives in depth-to-video synthesis task.}
    \label{fig:loss-ablation}
\end{figure}

\subsection{Ablation and analysis}
\label{sec:experiments_ablation}
\paragraph{Training objectives}
We compare three training objectives to validate our proposed stabilized velocity matching objective defined in Eq.~\eqref{eq:stabilized-velocity-loss}, along with displacement matching Eq.~\eqref{eq:displacement-loss} and velocity matching Eq.~\eqref{eq:vm-loss}.
Table~\ref{tab:ablation-objective} shows that stabilized velocity matching consistently achieves the best performance on both depth-to-video and image editing tasks. 
Specifically, it surpasses other objectives on all evaluated metrics for depth-to-video generation, and it also attains the highest average scores in diverse image editing scenarios.
Moreover, Figure~\ref{fig:loss-ablation} highlights its superior training stability and improved visual quality compared to alternative objectives.

\paragraph{Noise scale}
Several previous works~\cite{li2023bbdm, chadebec2025lbm} further extend the Brownian Bridge formulation by modifying the diffusion term in Eq.~\eqref{eq:general-sde}. 
Instead of fixing the diffusion coefficient as a constant $\sigma(t)\equiv1$, they introduce a global noise scale parameter $s$ such that $\sigma(t)\equiv s$, leading to the generalized SDE:
\begin{equation}
    \mathrm{d}X_t = v_\theta(X_t, t)\,\mathrm{d}t + s\,\mathrm{d}W_t.
    \label{eq:generalized-sde}
\end{equation}
To investigate its impact, we conduct experiments across different values of $s$, summarizing results in Table~\ref{tab:noise-scale}.
The corresponding training and inference modifications for this generalized formulation are detailed in the Supplementary Material, Section~\ref{sec:theoretical-derivations}.
Our findings indicate that moderate noise scales ($s=1$ or $s=2$) achieve better performance for the depth-to-video task, with $s=2$ showing strong overall scores.
For image editing tasks, a smaller noise scale ($s=0.5$) surprisingly achieves the highest average performance, notably outperforming the default $s=1$ setting.
However, excessively small ($s<0.5$) or large ($s>2$) noise scales significantly degrade quality on both tasks.
These observations highlight that optimal noise scales differ across tasks, contrasting with previous work~\cite{chadebec2025lbm} advocating an extremely small noise scale ($s=0.005$).

\section{Conclusion}
In this paper, we introduced the Visual Bridge Transformer, a large-scale instantiation of Brownian Bridge models, effectively scaling this paradigm to 20B parameters for conditional image and video generation.
By proposing a stabilized velocity-matching objective, we addressed the numerical instability inherent in conventional training methods, significantly improving model stability and performance.
Extensive experiments demonstrated that our framework consistently outperforms existing baselines across multiple challenging vision translation tasks, including instruction-based image editing and video translation tasks. 

\section{Limitations and future work}
While our Visual Bridge Transformer demonstrates strong results, we observed that adjusting the noise scale $s$ can further optimize performance across different vision tasks. Future work may explore adaptive or automated methods to select this parameter, potentially enhancing the versatility and effectiveness of Bridge Models.

{
\clearpage
\maketitlesupplementary

\appendix
\renewcommand{\thesection}{\Alph{section}}

\setcounter{table}{0}
\renewcommand{\thetable}{S\arabic{table}}

\setcounter{figure}{0}
\renewcommand{\thefigure}{S\arabic{figure}}

\setcounter{algocf}{0}
\renewcommand{\thealgocf}{S\arabic{algocf}}

\section{Additional experimental results}
\label{sec:additional_results}

\paragraph{Efficiency comparison}
The Brownian Bridge formulation in ViBT enables more efficient training and inference by reducing reliance on auxiliary conditional branches or additional conditioning tokens.
To quantitatively illustrate this potential advantage, we perform theoretical inference latency comparisons between ViBT and conventional conditional diffusion transformer (DiT) variants, which inject conditions by introducing extra tokens into attention layers.
For image translation, ViBT is instantiated from Qwen-Image-Editing~\cite{wu2025qwen}, while for video translation, ViBT is built upon Wan 2.1 1.3B~\cite{wan2025wan}.
Corresponding conditional DiT variants derived from these models serve as our baselines.
We measure the inference latency for a single forward pass under a single NVIDIA H200 GPU, ensuring a clean architectural efficiency comparison independent from sampling schedules or runtime optimization.

Tables~\ref{tab:efficiency_image} and~\ref{tab:efficiency_video} detail the raw data for this comparison, including exact token counts and per-step latencies under various image resolutions and video settings.
Figure~\ref{fig:latency-comparison} further visualizes the latency comparisons, clearly demonstrating that ViBT consistently reduces inference latency across all evaluated image and video translation scenarios compared to the conditional DiT baselines.

\paragraph{Additional video translation tasks}
Besides the depth-to-video synthesis task presented in Section~\ref{sec:experiments_translation}, we further evaluate ViBT on two additional video translation tasks: (1) {\textit{video colorization}} and (2) {\textit{video frame interpolation}}.

For video colorization, we directly apply ViBT to transform grayscale videos into colored videos. Figure~\ref{fig:video-colorization} shows qualitative examples of video colorization results, highlighting ViBT's strong generalization capability.

For video frame interpolation, we first construct a coarse video by repeating each original frame (except the first frame) $k$ times in pixel space.
ViBT is then applied to refine this coarse video, enhancing both visual quality and temporal coherence. Figure~\ref{fig:video-interpolation-pipe} illustrates this interpolation pipeline clearly.
In our experiments, we set $k=4$ to generate $4\times$ interpolated frames between each original frame. 
This increases the frame rate of videos generated by Wan 2.1 from 15 FPS to 60 FPS, while maintaining high visual quality and temporal coherence.
Qualitative results for this interpolation task are provided in Figure~\ref{fig:video-interpolation}.

Notably, ViBT is capable of producing high-quality and temporally coherent results within only a few inference steps (e.g., 4 steps), demonstrating its efficiency.

\begin{table}[t]
\centering
\scriptsize
\resizebox{\columnwidth}{!}{
\begin{tabular}{l|cc|ccc}
\toprule
\multirow{2}{*}{\textbf{Resolution}} & \multicolumn{2}{c|}{\textbf{Conditional DiT}} & \multicolumn{3}{c}{\textbf{ViBT}} \\
\cmidrule(lr){2-3} \cmidrule(lr){4-6}
& Tokens & Latency (ms) & Tokens & Latency (ms) & Speedup \\
\midrule
1024 $\times$ 1024 & 8,192 & 437 & 4,096 & 192 & 2.28$\times$ \\
1328 $\times$ 1328 & 10,624 & 613 & 5,312 & 258 & 2.38$\times$ \\
\bottomrule
\end{tabular}
}
\caption{Inference efficiency comparison (image).}
\label{tab:efficiency_image}
\end{table}

\begin{table}[t]
\centering
\scriptsize
\resizebox{\columnwidth}{!}{
\begin{tabular}{l|cc|ccc}
\toprule
\multirow{2}{*}{\textbf{Resolution}} & \multicolumn{2}{c|}{\textbf{Conditional DiT}} & \multicolumn{3}{c}{\textbf{ViBT}} \\
\cmidrule(lr){2-3} \cmidrule(lr){4-6}
& Tokens & Latency (ms) & Tokens & Latency (ms) & Speedup \\
\midrule
480P (5s) & 65,520 & 1,510 & 32,760 & 459 & 3.29$\times$ \\
480P (10s) & 127,920 & 5,407 & 63,960 & 1,444 & 3.74$\times$ \\
720P (5s) & 151,200 & 7,437 & 75,600 & 1,958 & 3.80$\times$ \\
720P (10s) & 295,200 & 28,577 & 147,600 & 7,097 & 4.03$\times$ \\
\bottomrule
\end{tabular}
}
\caption{Inference efficiency comparison (video).}
\label{tab:efficiency_video}
\end{table}

\begin{figure}[t]
    \centering
    \includegraphics[width=\linewidth]{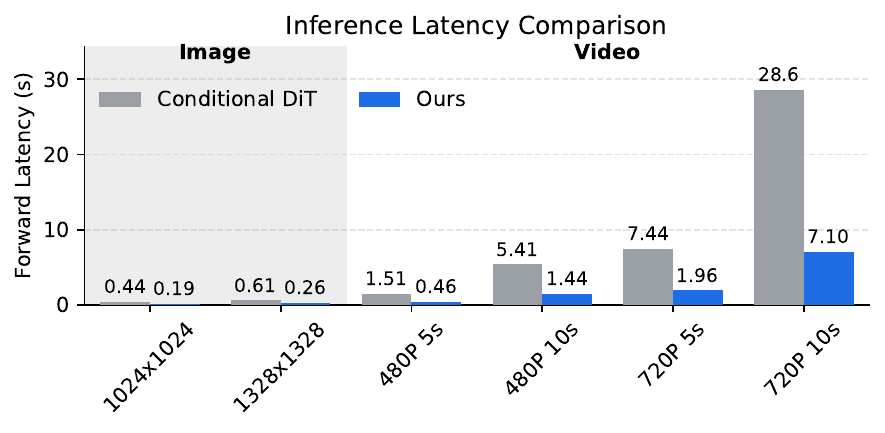}
    \vspace{-2.5em}
    \caption{Comparison between Conditional DiT and ViBT.}
    \label{fig:latency-comparison}
\end{figure}

\begin{figure}[t]
    \centering
    \includegraphics[width=\linewidth]{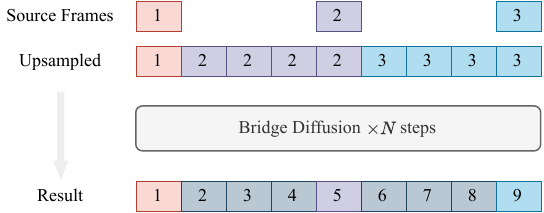}
    \caption{Illustration of video frame interpolation pipeline.}
    \label{fig:video-interpolation-pipe}
\end{figure}

\begin{figure*}[t]
    \centering
    \includegraphics[width=\textwidth]{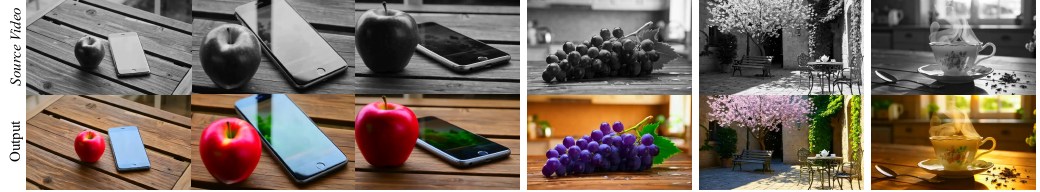}
    \caption{Qualitative results on video colorization task.}
    \label{fig:video-colorization}
\end{figure*}

\begin{figure*}[t]
    \centering
    \includegraphics[width=\textwidth]{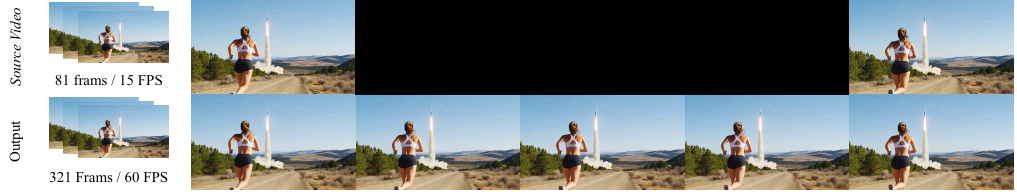}
    \caption{Qualitative results on video frame interpolation task.}
    \label{fig:video-interpolation}
    \vspace{-1em}
\end{figure*}

\paragraph{Ablation study on variance-corrected sampling}
To validate the effectiveness variance-corrected sampling strategy described in Eq.~\eqref{eq:stoch-update}, we perform an ablation study by comparing it with the standard Euler-Maruyama discretization method without variance correction.
Figure~\ref{fig:variance-corrected-sampling} provides qualitative results for this comparison on the image editing task.
We observe that the naive discretization method (without variance correction) introduces noticeable artifacts, leading to degraded visual quality. In contrast, the variance-corrected sampling generates a cleaner and visually coherent image.

\begin{figure}[h]
    \centering
    \includegraphics[width=\linewidth]{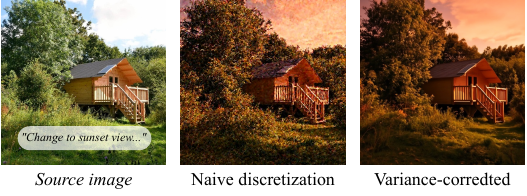}
    \caption{Ablation study on variance-corrected sampling.}
    \label{fig:variance-corrected-sampling}
\end{figure}

\paragraph{Influence of inference steps and schedule}
We further investigate how the number of inference steps and the discretization schedule affect ViBT's performance.
As illustrated in Figure~\ref{fig:inference-steps-scheduler}, increasing the inference steps consistently improves the generation quality.
Moreover, the choice of timestep scheduler significantly influences the performance.
Specifically, we adopt the shifting strategy introduced in \textit{Stable Diffusion 3}~\cite{esser2024scaling}, which uses a shift coefficient $\gamma$ to allocate more inference steps towards the earlier stages ($t \rightarrow 0$) of the diffusion process.
This shifted schedule is formulated as:
\begin{equation}
t_i = \frac{i}{\gamma\,N + (\gamma - 1)\, i},
\end{equation}
where $N$ denotes total steps and $i$ the step index.

Figure~\ref{fig:shifted-scheduler} illustrates how increasing $\gamma$ redistributes step density, placing more steps at earlier stages.
Our experiments show that $\gamma=5$ achieves significantly better visual quality than the linear schedule ($\gamma=1$), especially with fewer inference steps (e.g., 4 or 8 steps).

\begin{figure}[t]
    \centering
    \includegraphics[width=0.95\linewidth]{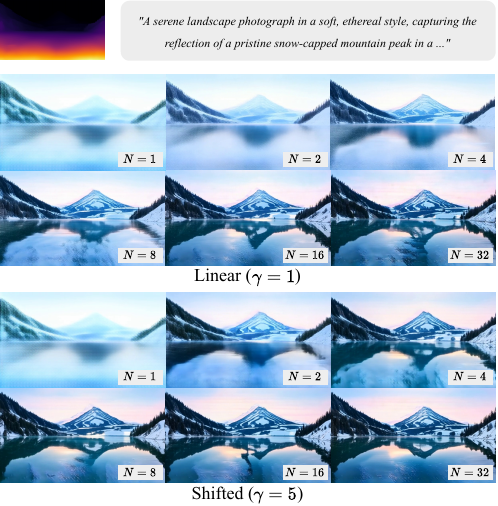}
    \vspace{-0.5em}
    \caption{Ablation on inference steps and timestep schedule.}
    \label{fig:inference-steps-scheduler}
\end{figure}

\begin{figure}[t]
    \centering
    \includegraphics[width=\linewidth]{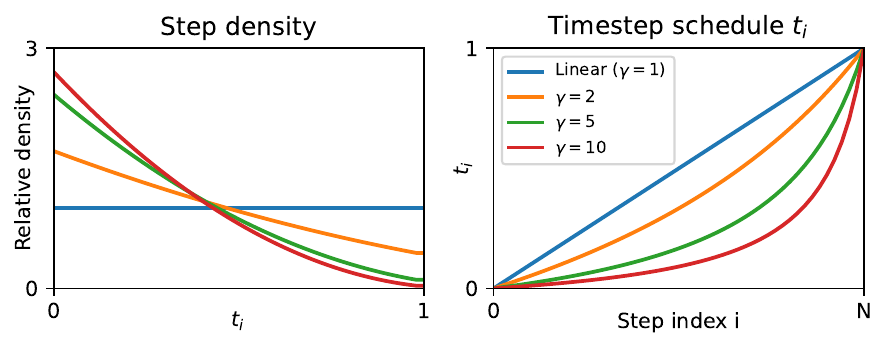}
    \vspace{-2em}
    \caption{Step density and timestep schedule for different $\gamma$.}
    \label{fig:shifted-scheduler}
\end{figure}

\begin{figure*}[t]
    \centering
    \includegraphics[width=\textwidth]{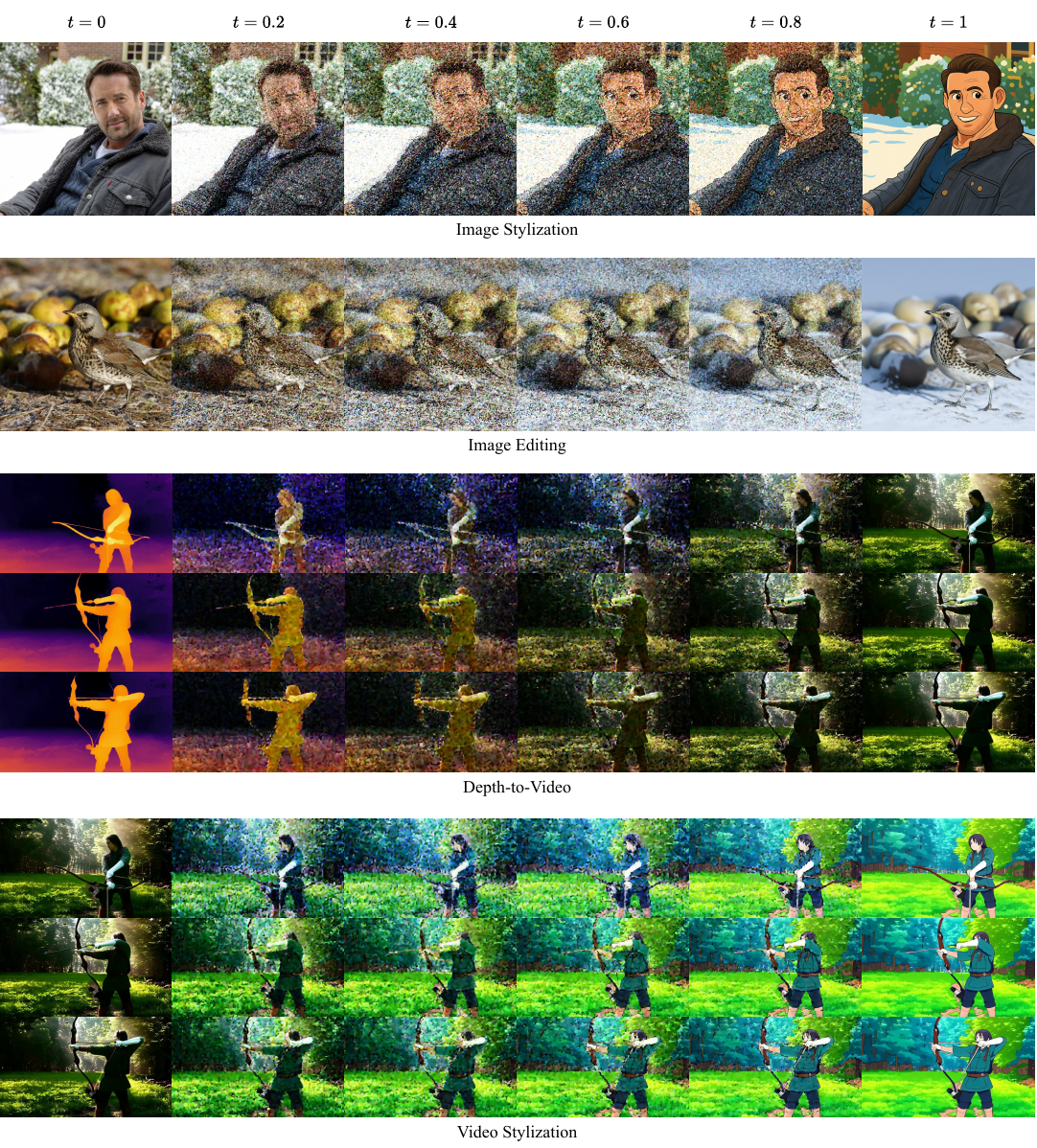}
    \caption{Visualization of the intermediate stages in the ViBT bridge process.}
    \label{fig:supp-bridge-process}
\end{figure*}

\paragraph{Additional visualizations}
We provide supplementary qualitative results: Figure~\ref{fig:supp-bridge-process} visualizes intermediate generation states at different timesteps $t$ in the ViBT bridge process, Figure~\ref{fig:supp-img-style} shows additional examples of image stylization tasks, Figure~\ref{fig:supp-img-edit} presents further results on instruction-based image editing, and Figure~\ref{fig:supp-video-style} provides extra visualizations of video stylization outputs.

\begin{figure*}[t]
    \centering
    \includegraphics[width=\textwidth]{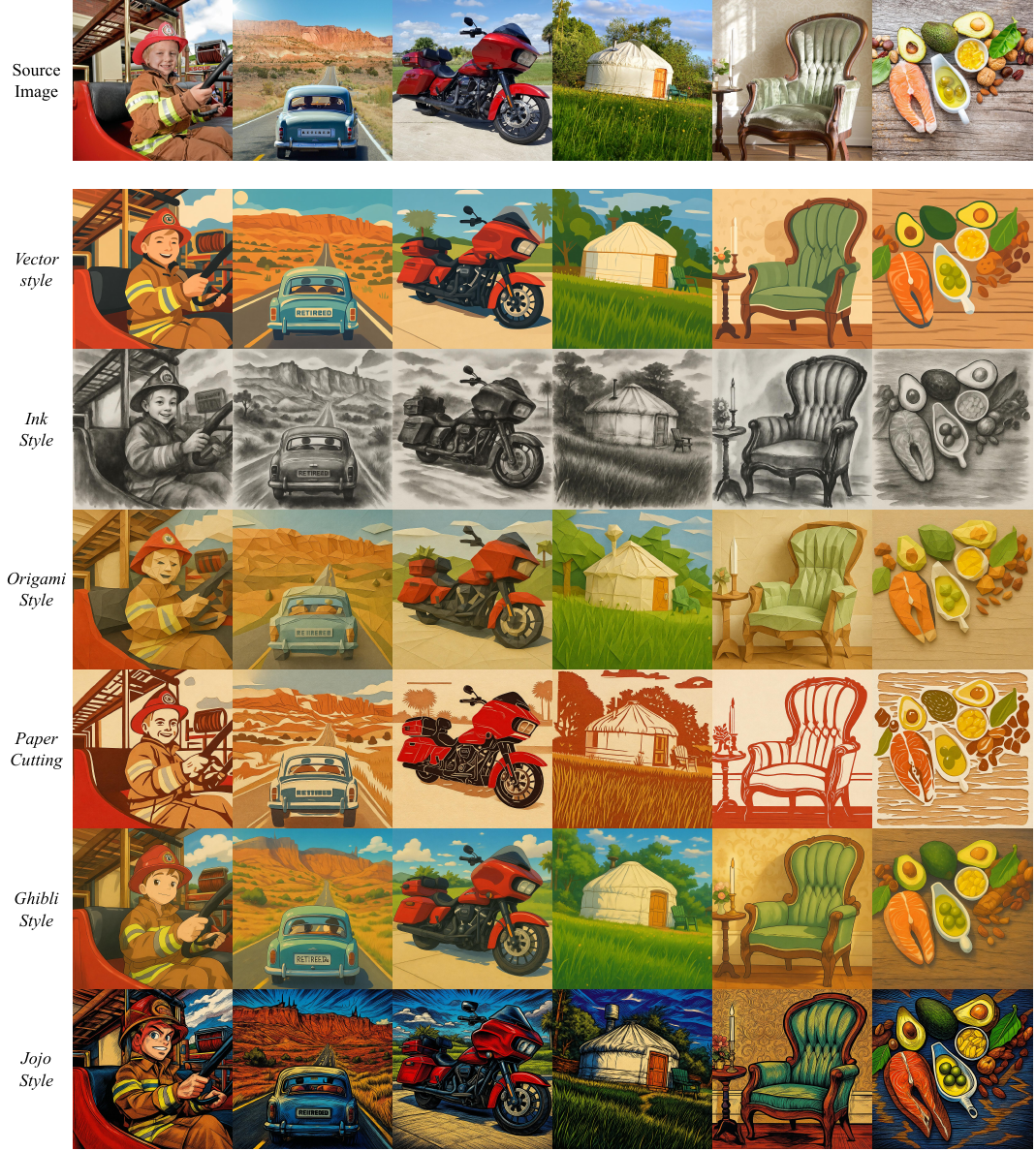}
    \caption{Additional examples of image stylization generated by ViBT.}
    \label{fig:supp-img-style}
\end{figure*}

\begin{figure*}[t]
    \centering
    \includegraphics[width=\textwidth]{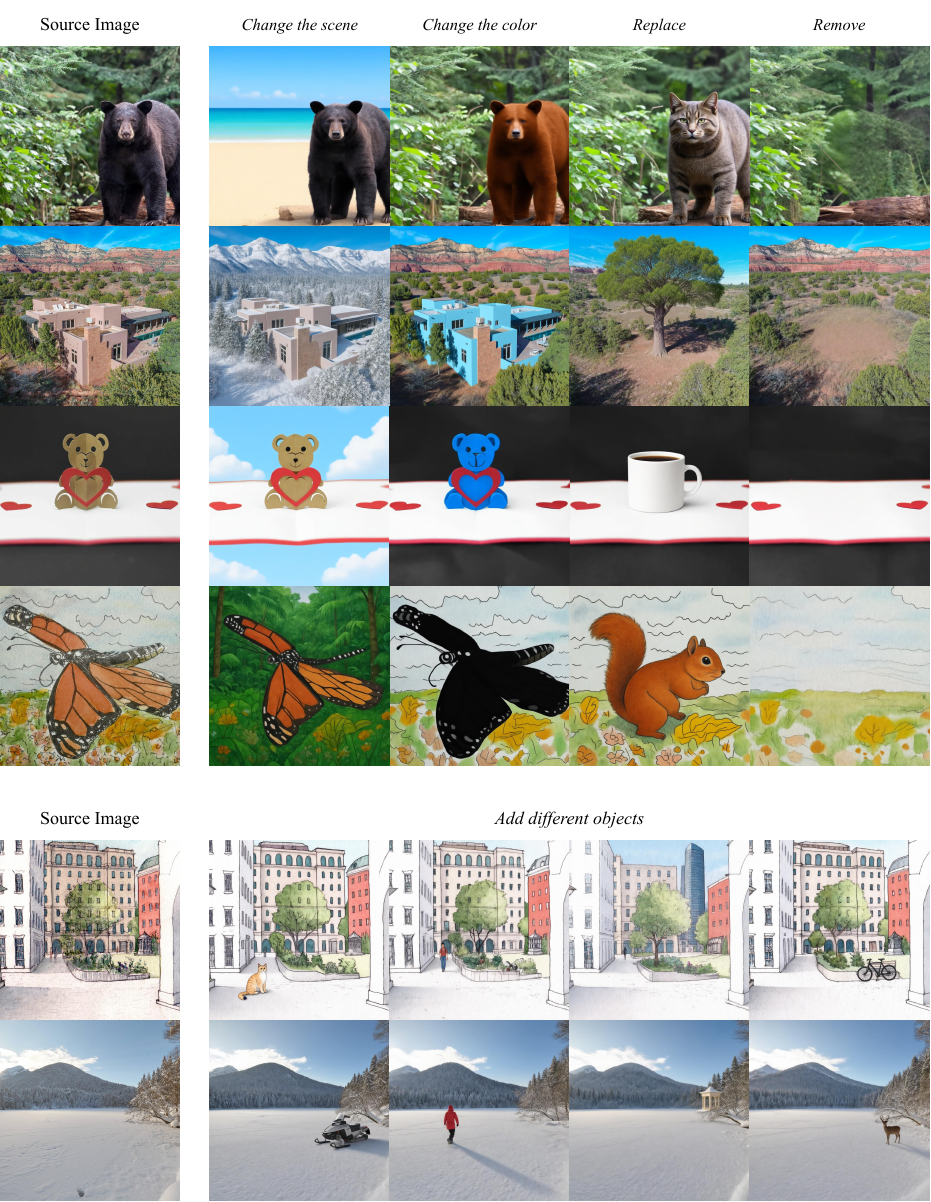}
    \caption{Additional qualitative results on instruction-based image editing.}
    \label{fig:supp-img-edit}
\end{figure*}

\begin{figure*}[t]
    \centering
    \includegraphics[width=\textwidth]{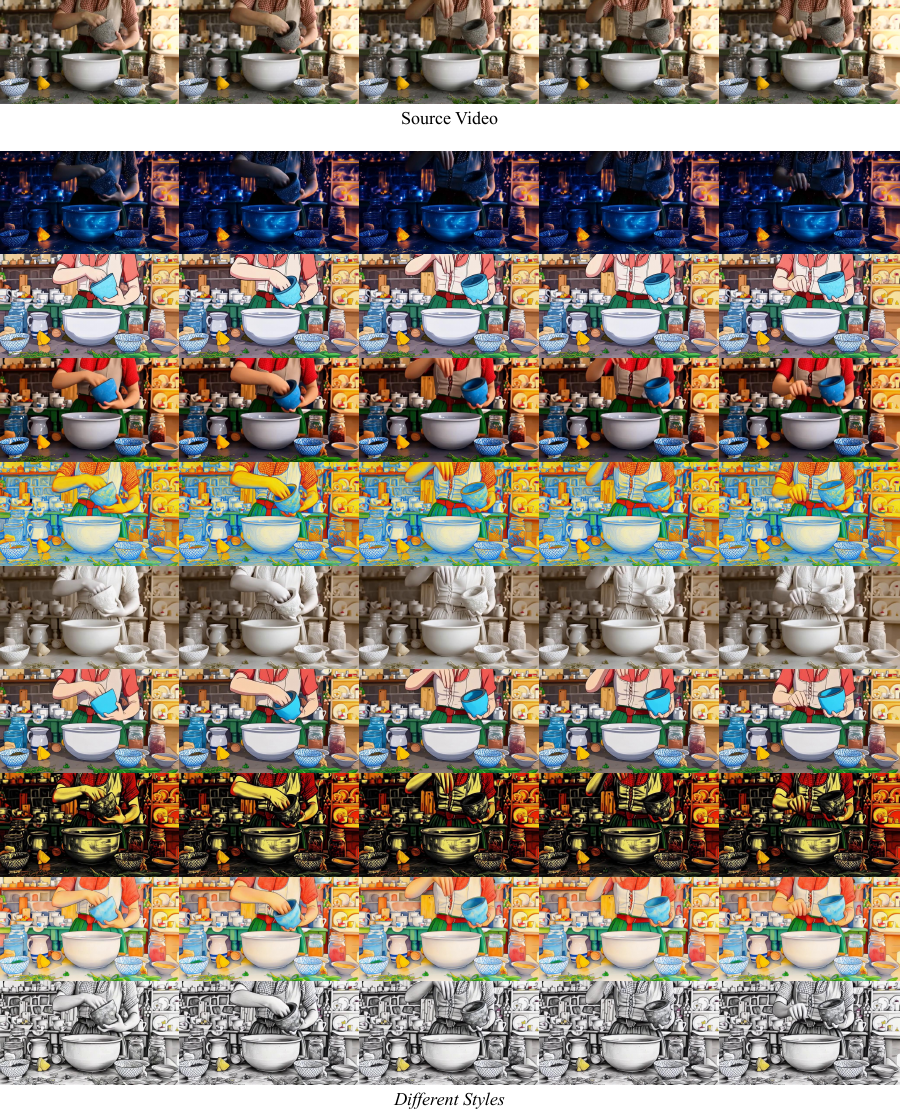}
    \caption{Additional results of video stylization tasks.}
    \label{fig:supp-video-style}
\end{figure*}
\clearpage
\clearpage
\section{Experimental details}
\label{sec:exp_detail}
\paragraph{Image editing dataset construction}
We construct our image editing dataset by first randomly sampling 5,000 images from the Open Images Dataset~\cite{kuznetsova2020open}. These images are cropped and resized into resolutions supported by the Qwen-Image-Editing model, specifically: $1328\times1328$ (1:1), $1664\times928$ (16:9), $928\times1664$ (9:16), $1472\times1104$ (4:3), $1104\times1472$ (3:4), $1584\times1056$ (3:2), and $1056\times1584$ (2:3). 
Subsequently, we generate corresponding editing instructions for these images using the vision-language model Qwen3-VL~\cite{yang2025qwen3}. We then produce edited images based on these instructions using Qwen-Image-Editing~\cite{wu2025qwen}. To ensure high-quality alignment, we further score the generated instruction-image pairs using Qwen3-VL, filtering out pairs with low alignment scores. This filtered set constitutes Part 1 of our training data, comprising approximately 3,335 validated samples.

Additionally, we enrich the dataset by incorporating stylized images generated by OmniConsistency~\cite{song2025omniconsistency}. These images retain their original $1024\times1024$ resolution, with editing instructions uniformly formulated as ``Convert the image to a [style] style image.'' This augmentation forms Part 2 of our dataset, introducing further diversity with approximately 2,605 samples.

\paragraph{Depth-to-Video dataset construction}
To create the training dataset for depth-to-video synthesis, we first generate 1,003 videos using Wan 2.2 14B~\cite{wan2025wan} with prompts sourced from the \textit{MovieGen Bench}~\cite{polyak2024movie}.
These videos are synthesized at a resolution of $832\times480$ with 81 frames each, using a classifier-free guidance (CFG) scale of 5 and 50 sampling steps.

We then transform these synthesized videos into depth maps using the \textit{Depth Anything V2}~\cite{depth_anything_v2} model, forming depth-video pairs for training.
It should be noted that the generated depth maps utilize the default \texttt{inferno} colormap format provided by Depth Anything V2, rather than grayscale images.

\paragraph{Depth-to-Video evaluation details}
For evaluation on the depth-to-video synthesis task, we generate 946 reference videos using Wan 2.2 14B based on the prompts provided by VBench~\cite{huang2023vbench}.
These videos are also synthesized at a resolution of $832\times480$ with 81 frames each, using a CFG scale of 5 and 50 sampling steps.
We then convert these videos into corresponding depth maps using the \textit{Depth Anything V2}~\cite{depth_anything_v2} model, which are employed as conditioning inputs across all methods.
The prompts used for generating the source videos are the extended versions~\cite{wan2025wan}.
However, for fair evaluation during testing, we use the original prompts provided by VBench.
\section{Theoretical analysis and extensions}
\label{sec:theoretical-derivations}

\paragraph{Normalization factor for stabilized velocity matching}
Conditioned on endpoints $(x_0,x_1)$, the Brownian Bridge latent at time $t$ can be written as
\begin{equation}
    x_t
    = (1 - t)x_0 + t x_1
      + \sqrt{t(1 - t)}\,\epsilon,
    \qquad \epsilon \sim \mathcal{N}(0, I).
\end{equation}
The velocity target is
\begin{equation}
    u_t(x_t \mid x_1)
    = \frac{x_1 - x_t}{1 - t}.
\end{equation}
Substituting $x_t$ gives
\begin{align}
    x_1 - x_t
    &= (1 - t)(x_1 - x_0)
       - \sqrt{t(1 - t)}\,\epsilon, \\
    u_t(x_t \mid x_1)
    &= (x_1 - x_0)
       - \sqrt{\frac{t}{1 - t}}\,\epsilon.
\end{align}

We define the normalization factor via the (conditional) expected squared normlknjugytfrd5e4s3w2aq1 fc gvbhnjm
\begin{equation}
    \alpha(x_0,x_1,t)^2
    =
    \frac{\mathbb{E}_\epsilon
        \bigl[\|u_t(x_t \mid x_1)\|^2\bigr]}
         {\|x_1 - x_0\|^2},
\end{equation}
where the expectation is taken over $\epsilon$ with $(x_0,x_1)$ fixed.
Using $\mathbb{E}[\epsilon]=0$ and $\mathbb{E}[\|\epsilon\|^2]=D$, we obtain
\begin{align}
    \mathbb{E}_\epsilon
    \bigl[\|u_t(x_t \mid x_1)\|^2\bigr]
    &= \bigl\|x_1 - x_0\bigr\|^2
     + \frac{t}{1 - t}\,D,
\end{align}
and hence
\begin{align}
    \alpha(x_0,x_1,t)^2
    &= \frac{\|x_1 - x_0\|^2
            + \tfrac{t}{1 - t}D}
           {\|x_1 - x_0\|^2} \\
    &= 1
     + \frac{t\,D}
            {(1 - t)\,\|x_1 - x_0\|^2}.
\end{align}
This is the normalization factor used in the stabilized velocity loss in Eq.~\eqref{eq:stabilized-velocity-loss}.

\paragraph{Variance-corrected noise scaling}

Write the process as a deterministic interpolation plus a zero-mean Brownian Bridge:
\begin{equation}
    X_t
    = (1 - t)x_0 + t x_1 + B_t,
\end{equation}
where $\{B_t\}_{t\in[0,1]}$ is a Brownian Bridge from $0$ to $0$.
For $0 \le t_1 \le t_2 \le 1$, its covariance satisfies
\begin{align}
    \mathbb{E}[B_{t_2}] &= 0, \\
    \operatorname{Var}(B_{t_2}) &= t_2(1 - t_2)\,I, \\
    \operatorname{Cov}(B_{t_1}, B_{t_2})
        &= t_1(1 - t_2)\,I.
\end{align}
Thus the conditional variance is
\begin{align}
    \operatorname{Var}(B_{t_2} \mid B_{t_1})
    &= \operatorname{Var}(B_{t_2}) \\
    &\quad
       - \operatorname{Cov}(B_{t_2},B_{t_1})\,
         \operatorname{Var}(B_{t_1})^{-1} \\
    &\quad\quad
       \cdot \operatorname{Cov}(B_{t_1},B_{t_2}) \\
    &= t_2(1 - t_2)\,I
     - \frac{t_1^2(1 - t_2)^2}{t_1(1 - t_1)}\,I \\
    &= \frac{(t_2 - t_1)(1 - t_2)}{1 - t_1}\,I.
\end{align}

Since the endpoints only affect the mean,
\begin{equation}
    \operatorname{Var}(X_{t_2} \mid X_{t_1})
    = \frac{(t_2 - t_1)(1 - t_2)}{1 - t_1}\,I.
\end{equation}

For a discretization schedule
\begin{equation}
    0 = t_0 < t_1 < \dots < t_N = 1,
\end{equation}
set $t_1 = t_k$, $t_2 = t_{k+1}$ and
$\Delta t_k = t_{k+1} - t_k$ to obtain
\begin{equation}
    \operatorname{Var}\bigl(X_{t_{k+1}} \mid X_{t_k}\bigr)
    = \Delta t_k\,
      \frac{1 - t_{k+1}}{1 - t_k}\,I.
\end{equation}
Therefore an increment of the form
\begin{equation}
    X_{t_{k+1}}
    = X_{t_k}
      + \sqrt{\Delta t_k\,
              \frac{1 - t_{k+1}}{1 - t_k}}\,\epsilon_k,
    \quad \epsilon_k \sim \mathcal{N}(0,I),
\end{equation}
matches the Brownian Bridge conditional variance.

When discretizing
\begin{equation}
    \mathrm{d}X_t
    = v_\theta(X_t,t)\,\mathrm{d}t
      + \mathrm{d}W_t,
\end{equation}
the Euler--Maruyama update with variance correction becomes
\begin{align}
    x_{k+1}^{\text{corrected}}
    &= x_k
     + \Delta t_k\,v_\theta(x_k,t_k) \nonumber\\
    &\quad
     + \sqrt{\Delta t_k\,
             \frac{1 - t_{k+1}}{1 - t_k}}\,\epsilon_k,
\end{align}
which is precisely the update in Eq.~\eqref{eq:stoch-update}.

\paragraph{Training and inference with noise scale}
Under the generalized Brownian Bridge SDE in Eq.~\eqref{eq:generalized-sde}, we keep the network architecture and stabilized velocity objective unchanged, and only rescale the stochastic terms by the global noise scale $s$.
Concretely, the intermediate state construction in training and the variance-corrected noise in inference are both multiplied by $s$, as summarized below.

\begin{algorithm}[h]
    \caption{Training with noise scale $s$}
    \label{alg:training-noise-scale}
    \KwIn{data pairs $(x_0,x_1)\sim p_{\text{source,target}}$, model $v_\theta$, latent dimension $D$, noise scale $s$}
    \Repeat{convergence}{
        Sample latent pair $(x_0,x_1)$, interpolation time $t\sim U(0,1)$, and noise $\epsilon\sim\mathcal{N}(0,I)$\;
        Construct intermediate state $x_t = (1-t)x_0 + t x_1 + s\sqrt{t(1-t)}\,\epsilon$\;
        Compute velocity target $u_t = (x_1 - x_t)/(1-t)$\;
        Compute normalization factor $\alpha^2 = 1 + {s^2 tD}/{[(1 - t)\|x_1 - x_0\|^2]}$\;
        Compute stabilized velocity loss $\mathcal{L}_{\tilde{\text{velocity}}} = \|\frac{v_\theta(x_t,t)-u_t}{\alpha}\|^2$\;
        Update model parameters $\theta$ by gradient descent on $\mathcal{L}_{\tilde{\text{velocity}}}$\;
    }
\end{algorithm}

\begin{algorithm}[h]
    \caption{Inference with noise scale $s$}
    \label{alg:inference-noise-scale}
    \KwIn{source-target latent pair $(x_0,x_1)$, trained model $v_\theta$, latent dimension $D$, discretization steps $N$, discretization schedule $0=t_0<t_1<\dots<t_N=1$, noise scale $s$}
    Initialize $x \gets x_0$\;
    \For{$k=0,1,\dots,N-1$}{
        Compute step size $\Delta t \gets t_{k+1}-t_k$\;
        Compute scaling factor $\eta \gets s\sqrt{\Delta t \frac{1 - t_{k+1}}{1 - t_k}}$\;
        Sample noise $\epsilon\sim\mathcal{N}(0,I)$\;
        Update latent state:
        \[x \gets x + \Delta t\,v_\theta(x, t_k) + \eta\,\epsilon\]
    }
    \KwOut{Final state $x$ approximating the target $x_1$}
\end{algorithm}
}
\clearpage
\section*{Acknowledgment}
This project is supported by NUS IT’s Research Computing group under grant number NUSREC-HPC-00001.
We thank Ruonan Yu and Sicheng Feng for helpful discussions.

{
    \small
    \bibliographystyle{ieeenat_fullname}
    \bibliography{main}
}
\end{document}